\documentclass[
  english,        
  font=times,     
  twocolumn,      
]{tumarticle}

\usepackage{lipsum}
\usepackage[
backend=biber,
sorting=ynt
]{biblatex}
\usepackage{graphicx}
\title{ROBOT SKILL LEARNING VIA CLASSICAL ROBOTICS-BASED GENERATED DATASETS: ADVANTAGES, DISADVANTAGES AND FUTURE IMPROVEMENT}
\subtitle{GitHub: https://github.com/btknzn/TiagoRobotSkillLearning}

\author[affil=1, email = ozenbatukaan@gmail.com]{Batu Kaan Oezen}

\date{January 15, 2023}

\begin{document}

\maketitle

\begin{abstract}
    \textit{Why do we not profit from our long-existing classical robotics knowledge and look for some alternative way for data collection? The situation ignoring all existing methods might be such a waste. This article argues that a dataset created using a classical robotics algorithm is a crucial part of future development. This developed classic algorithm has a perfect domain adaptation and generalization property, and most importantly, collecting datasets based on them is quite easy. It is well known that current robot skill-learning approaches perform exceptionally badly in the unseen domain, and their performance against adversarial attacks is quite limited as long as they do not have a very exclusive big dataset. Our experiment is the initial steps of using a dataset created by classical robotics codes. Our experiment investigated possible trajectory collection based on classical robotics. It addressed some advantages and disadvantages and pointed out other future development ideas.}

\textbf{Keywords: Robot skill learning, Imitation learning\cite{mandlekar2020human}, Data collection strategies, MoveIt framework\cite{chitta2016moveit} , Tiago robot\cite{pages2016tiago}, Ros Navigation Stack\cite{guimaraes2016ros}.}
\end{abstract}

\section{Introduction}

Current robot skill learning algorithms are either based on smaller and similar datasets or massive and diverse trajectories datasets. Each of them has its own advantages and disadvantages. For example,  one basic robot skill  can be learned by just overfitting on one small dataset, but this situation will cause bad generalization. Robot skill learning tasks based on similar small datasets are generally overfitted in literature, and their performance is limited because of not diverse motion variety. The extensive dataset for robot skill learning is also limited, and they do not have various environments, robots, and cases. This situation blocks possible real-life usage of learning-based robotics. Many experiments\cite{burda2018large} show the importance of extensive datasets related to robot skill learning. But it is still an open question of the best way to collect robot skill learning datasets. Kinesthetic\cite{akgun2012trajectories} teaching and Teleoperation\cite{mandlekar2020human} are trendy in literature, but their inefficiency might prohibit them from being the correct way to collect datasets. On the other hand, reinforcement learning-based\cite{fayjie2018driverless} agents achieve amazing work, such as autonomous driving tasks. However, the convergence of reinforcement learning-based\cite{kumar2022should} agents is extremely long for robot skill learning tasks, and their real-life application might be dangerous.

Most importantly, researchers developed a lot of classical robotics algorithms over decades, and it is a big waste not using our solid robotics knowledge for  dataset collection. In literature, this way of dataset collection is limited or rare. This situation brings us to using classical robotics  for trajectory-based dataset collection, achieving robot skill learning tasks, and investigating our mentioned methods. 
During the experiment, robotics manipulation tasks are performed based on computer vision(object localization), slam(mapping and navigation), control theory, and robotics via using MoveIt\cite{chitta2016moveit} framework and ROS navigation stack\cite{zheng2021ros}. It impresses the advantages and disadvantages of the mentioned method for robot skill learning and points the way for future robot learning development.

\begin{figure}
    \includegraphics[width=\linewidth]{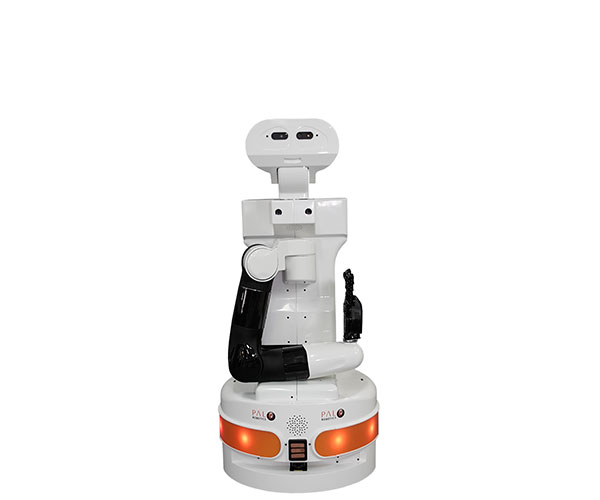}
    \caption{Tiago Robot\cite{pages2016tiago} }
    \label{fig:2.2}
\end{figure}

\section{Related Work}

    Collected datasets, learning algorithms, and neural network structure might be the most important part of robot skill learning tasks. This part discusses possible learning algorithms and neural network structures.

    When we come to the learning algorithm, there are two milestone learning algorithms for robot skill learning. One of them is imitation learning\cite{mandlekar2020human} gained a lot of attention because of its end-to-end learning success in robot skill learning, and the other one, Reinforcement learning\cite{kumar2022should}, started later showing some incredible success in the area, but it currently often overperforms imitation learning for robot skill learning. The imitation learning\cite{mandlekar2020human} approach performs similarly to classical deep neural network architecture. It only needs one robot manipulation task dataset and learns all possible actions using this dataset. On the other hand, some of the Reinforcement learning methods\cite{kumar2022should} are quite popular in robot skill learning.  Classical reinforcement learning\cite{kumar2022should} based on designed rewards,  inverse reinforcement learning\cite{hadfield2016cooperative}, and Offline reinforcement learning methods\cite{kumar2022should} are some of the popular reinforcement learning algorithms\cite{kumar2022should} for acquiring abilities for the robot. A short summary of these mentioned methods can be defined as upcoming sentences. 
    Inverse Reinforcement Learning (IRL)\cite{hadfield2016cooperative}is a method used in robotics to learn an agent's reward function from its observed behavior. It is an inverse problem of the traditional Reinforcement Learning\cite{fayjie2018driverless} (RL) approach, where an agent learns a policy by maximizing a known reward function. In IRL \cite{hadfield2016cooperative}, the goal is to infer the underlying reward function that generated a given set of expert demonstrations. Offline Reinforcement Learning\cite{kumar2022should} (ORL) is a method of training reinforcement learning\cite{kumar2022should} (RL) agents using previously collected data rather than online interactions with the environment. This approach is similar to imitation learning\cite{mandlekar2020human}, the only difference is that instead of training deep neural networks like imitation learning\cite{mandlekar2020human}, where the next action is predicted, there is a trained reinforcement learning\cite{kumar2022should} algorithm with collected trajectories. The experiment\cite{kumar2022should} defends that the offline reinforcement learning algorithms\cite{kumar2022should} performed better than the imitation learning\cite{mandlekar2020human} algorithm for robot skill learning tasks for some aspects. Handling partial observability and stochasticity, sub-optimal expert demonstration, generalization, and future improvement via online RL\cite{kumar2022should} are advantages over imitation learning\cite{mandlekar2020human}. However, the simplicity of imitation learning\cite{mandlekar2020human} made its usage more popular. It was our main reason for using imitation learning\cite{mandlekar2020human} in our task.

    Representation learning\cite{nguyen2021sensorimotor} and predictive learning\cite{nguyen2021sensorimotor} are one of the other most crucial crucial part for robot skill learning. Representation learning\cite{nguyen2021sensorimotor} compresses the information coming from sensors and produces a latent and sparse representation of inputs. This produced lower dimensional representation cause the system to learn with lower computation power and fewer parameters in the predictive learning\cite{nguyen2021sensorimotor} part. Generally, variational convolutional encoders\cite{ichiwara2022contact} and convolutional autoencoders\cite{ichiwara2022contact} are the most popular mechanism. The variational encoder\cite{ichiwara2022contact} has one advantage over classical encoders\cite{ichiwara2022contact} because they produce generally normalized latent information, and it improves the performance of the predictive learning\cite{nguyen2021sensorimotor} part. Generally, LSTM\cite{ichiwara2022contact}, attention-LSTM \cite{cui2020self}, and transformers-based architectures\cite{dasari2021transformers} are preferred in the predictive part. The main goal of the predictive learning\cite{nguyen2021sensorimotor} part is developing one neural network that can guess robot state for one step later, and it can select appropriate action based on that knowledge.

    \begin{figure}
    \includegraphics[width=\linewidth]{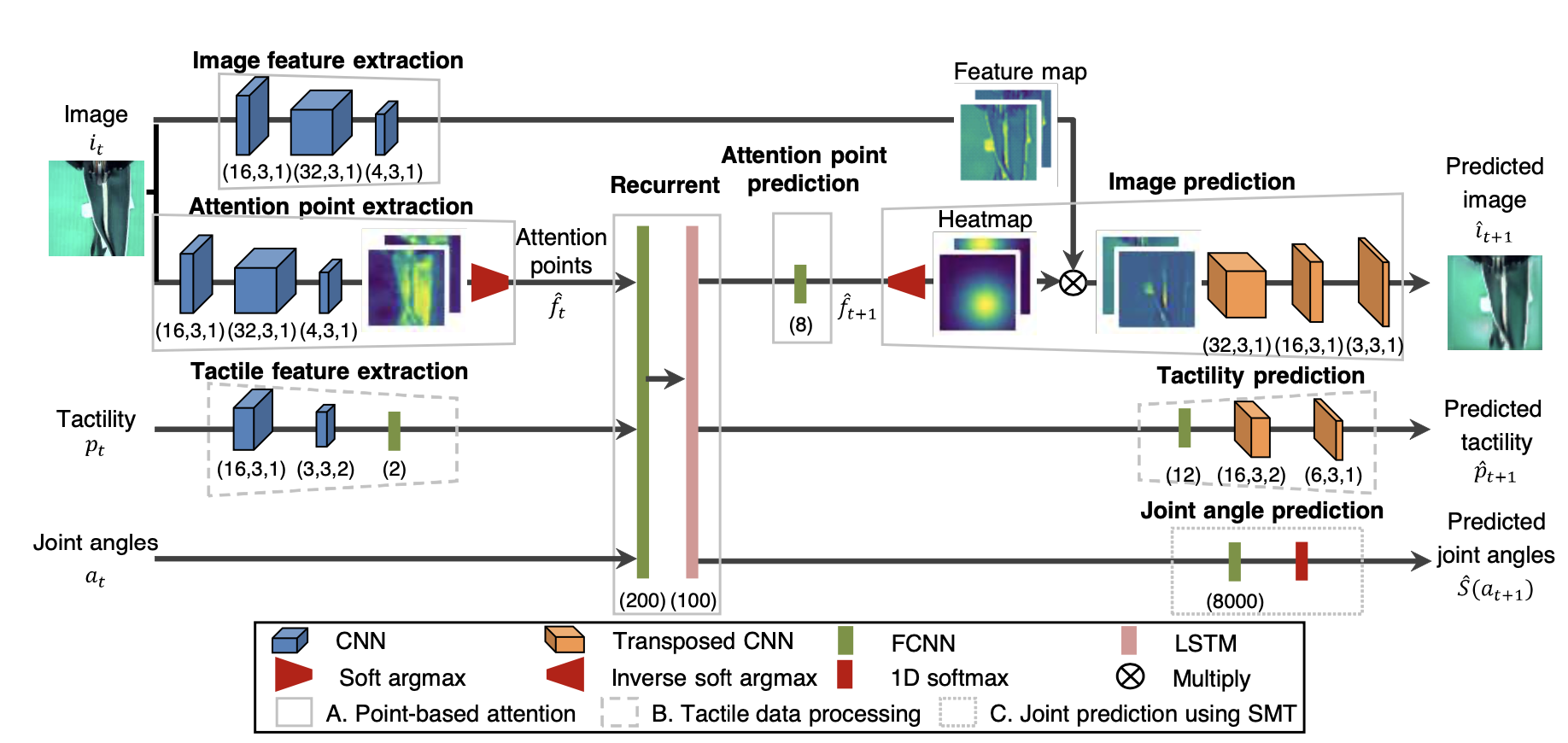}
    \caption{Example LSTM Neural network\cite{ichiwara2022contact} based on predictive learning and representation learning }
    \label{fig:2.2}
    \end{figure}
      \begin{figure}
    \includegraphics[width=\linewidth]{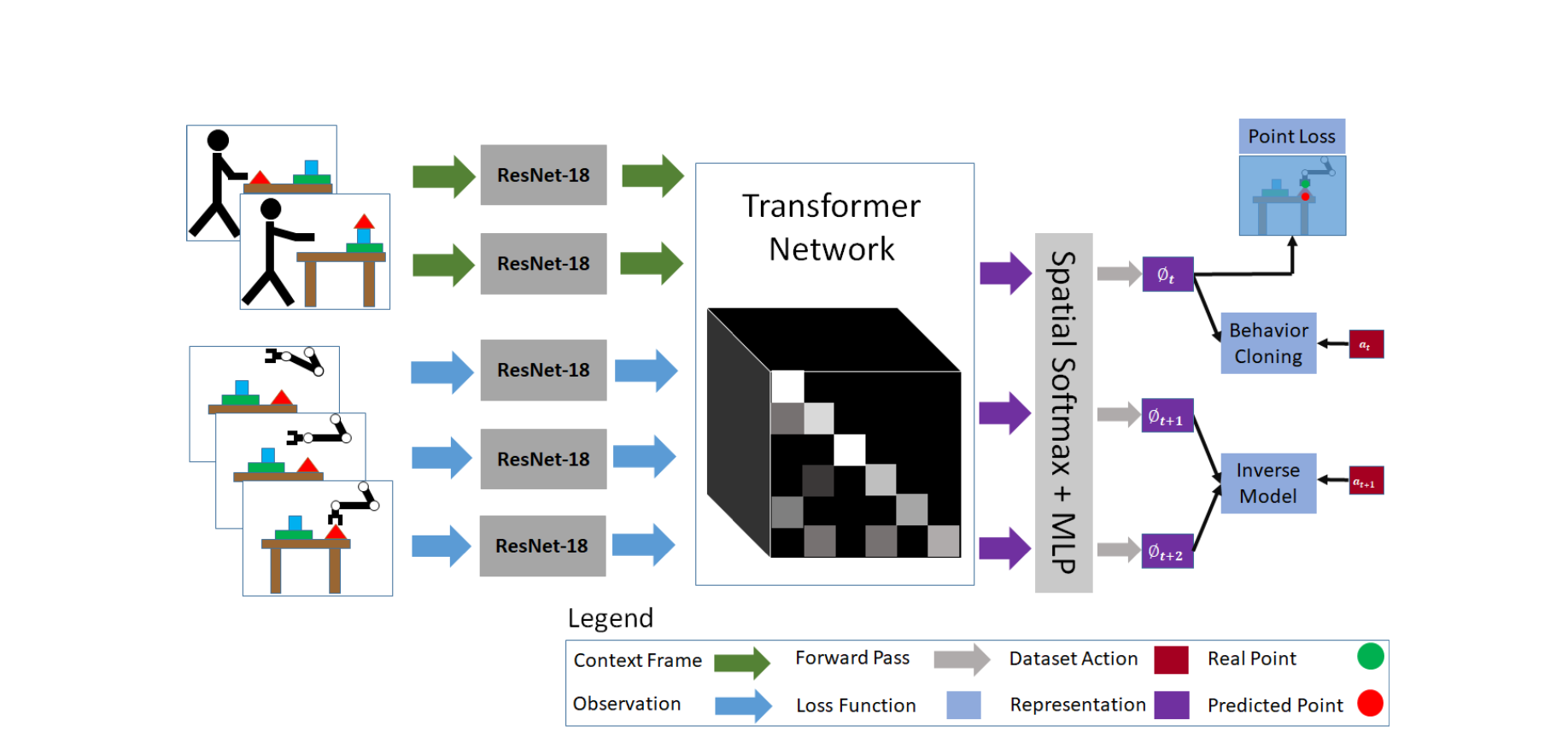}
    \caption{Example Transformer Neural network\cite{dasari2021transformers} based on predictive learning and representation learning }
    \label{fig:2.2}
    \end{figure}

    \begin{figure}
    \includegraphics[width=\linewidth]{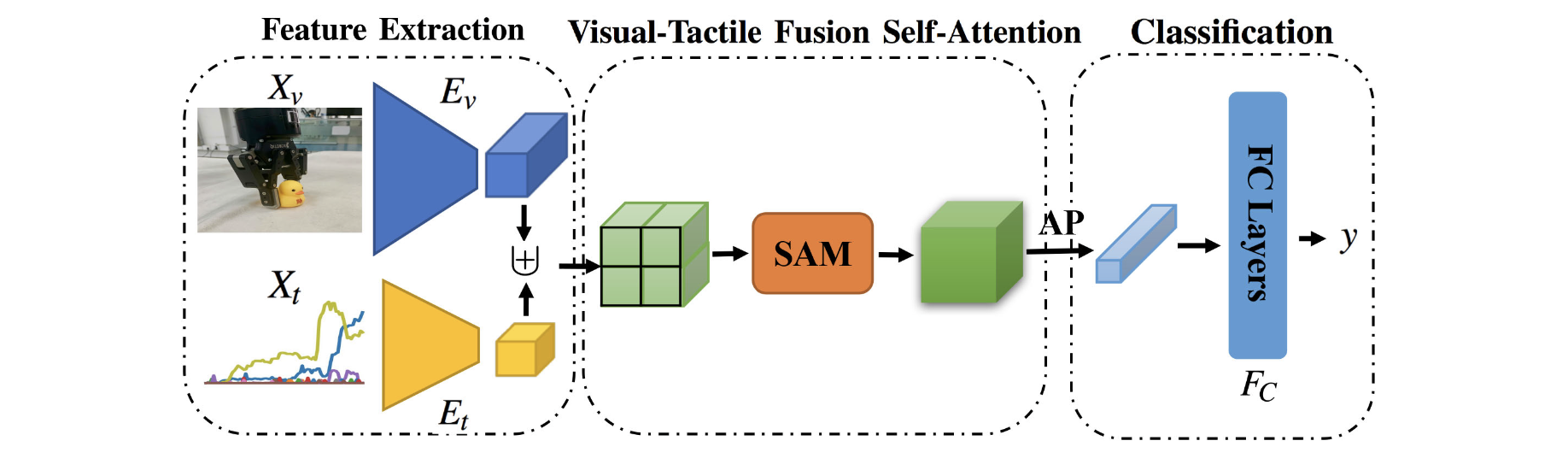}
    \caption{Example attention based Neural network\cite{cui2020self} based on predictive learning and representation learning }
    \label{fig:2.2}
    \end{figure}

\section{Dataset Collection}

This part uses two different dataset collection methods. The first one is for the long horizon, and the second one is for the short horizon.

Our first designed system uses a point cloud-based based object localization and Movebase\cite{zheng2021ros} based approaching to object. Movebase\cite{zheng2021ros} is a ROS (Robot Operating System) package that provides an implementation of a navigation stack\cite{zheng2021ros}. The navigation stack\cite{zheng2021ros} is a collection of algorithms that are used to plan and execute safe and efficient trajectories for a robot to move from one location to another. The specific algorithms used in the Movebase\cite{zheng2021ros}
package may vary depending on the particular implementation, but they typically include global and local planners, a path controller, and collision avoidance mechanisms.
The global planner algorithm is responsible for creating map based, while the local planner algorithm is responsible for generating detailed control commands for the robot to execute. The path controller is responsible for following the planned path, and the collision avoidance mechanisms are used to ensure that the robot does not collide with obstacles in its environment. For grasping the object, MoveIt\cite{chitta2016moveit}  is used, which is based on Inverse Kinematics, Forward Kinematics, collision detection, and path planning and following the planned path. In figure 2, you can see the designed GUI for the robot manipulation task. This  GUI is developed just for testing functions. Our object localization algorithm for point cloud is located in figure 7.

    \begin{figure}
    \includegraphics[width=\linewidth]{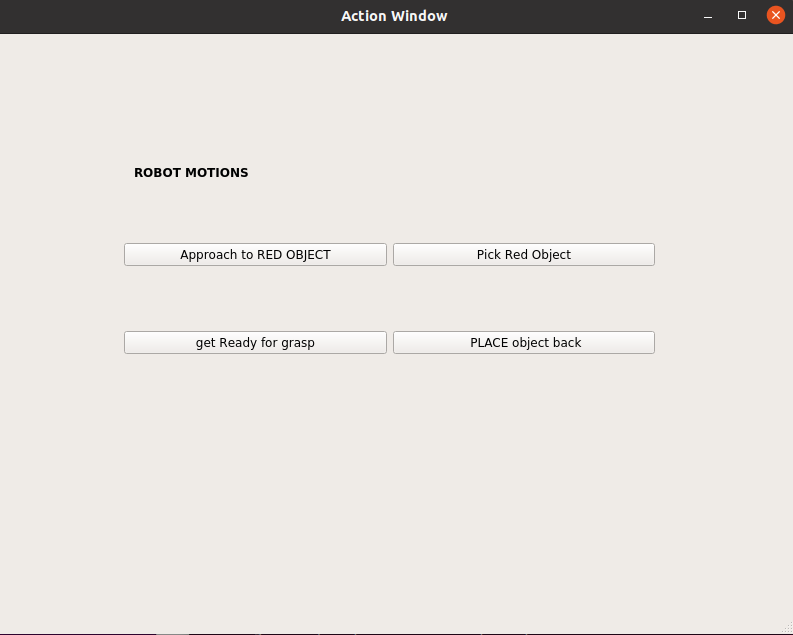}
    \caption{GUI for using Tiago Robot\cite{pages2016tiago} Robot }
    \label{fig:2.2}
    \end{figure}

    \begin{figure}
    \includegraphics[width=\linewidth]{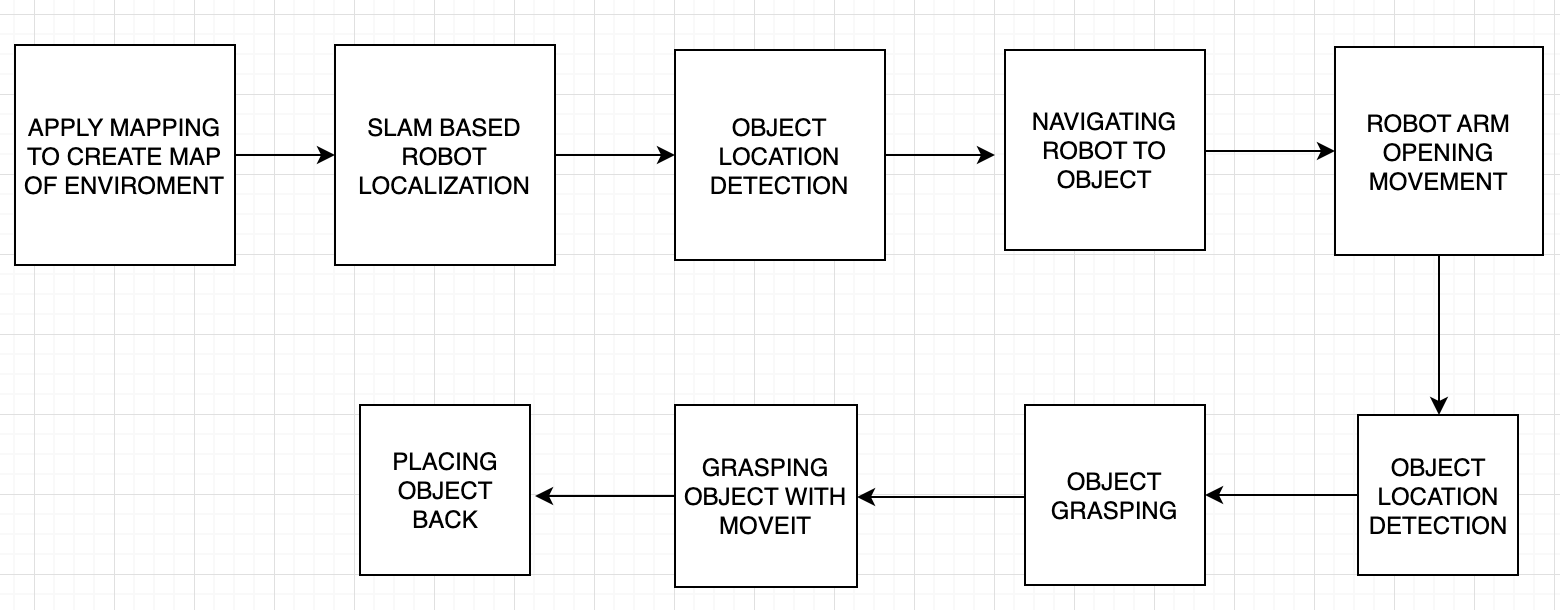}
    \caption{Action flow chart}
    \label{fig:2.2}
    \end{figure}

    \begin{figure}
    \includegraphics[width=\linewidth]{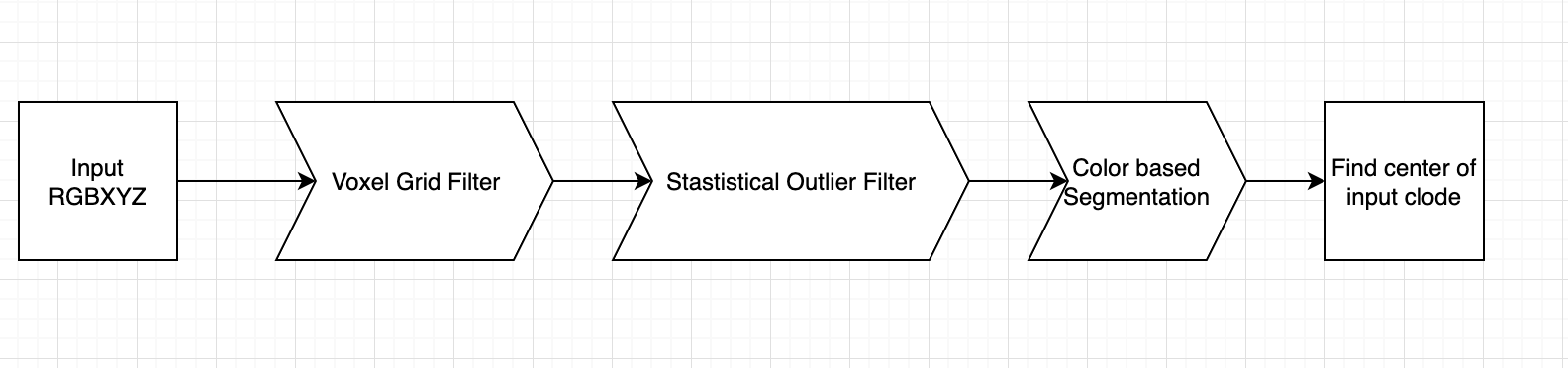}
    \caption{Object Localization Frame Work }
    \label{fig:2.2}
    \end{figure}

Figure 6 shows the action follow charts of our developed algorithm. Environmental mapping is completed via the navigation stack\cite{zheng2021ros}; afterward, robot localization is completed using created map features. The robot starts to localize the object like in figure 7; it uses a voxel grid filter\cite{xiong2021voxel} to reduce the input dataset to increase computation efficiency, then remove noise features via a statistical outlier filter\cite{balta2018fast}. Lastly, the object is localized color-based segmentation\cite{zhan2009color}. The Center location of the point cloud is the object location.

The second dataset collection method for robot object grasping was created based on Tiago\cite{pages2016tiago} pre-implemented by the producer company grasping code. This implementation uses ArUco marker\cite{lebedev2020accurate} based object localization using OpenCV\cite{bradski2000opencv}/PCL\cite{rusu20113d} library and grasping using MoveIt\cite{chitta2016moveit} library to grasp the object. During the recording, 10 different grasping objects are recorded for both implementations. The recorded dataset consists of all Tiago\cite{pages2016tiago} joints states, move base command, RGB image, and disparity image for the first described method. Another recording dataset has all \cite{pages2016tiago} joint states, RGB images, and disparity images.

\section{Skill Learning}

Imitation learning\cite{mandlekar2020human} is selected for analyzing our collected trajectories because of its simplicity. It uses the principle of using current information and predicting the next state, and the robot decides the next state using this method in real-time application. Below part, two different robotics software are used to collect for trajectory. When the dataset is investigated, the one with Movebase\cite{zheng2021ros}-MoveIt\cite{chitta2016moveit}  has complexity and a long recording time. It is also detected that MoveIt\cite{chitta2016moveit}  did not produce a similar grasping pattern for each grasping. 

The learning part has two main part. In the first part, two different encoder-decoder\cite{ichiwara2022contact}  is trained for RGB and disparity images on the Cifar-10 dataset\cite{recht2018cifar}. The second part, the architecture shown in figure 5, is trained end to end. Images and Disparities are standardized, and joint angles and move-base commands are normalized. The mean square loss function is defined as in equation 1. It uses a predictive learning\cite{nguyen2021sensorimotor} approach based on the current image frame, current robot state, and current disparity and predicts one step later robot state, based on this information, real-time robot application decides the next actions. The loss function calculates the MSE between next robot state and predicted next robot state.

$\mathcal{L}(x_{t+1}, \hat{x}{t+1}) = \frac{1}{N}\sum (x_{t+1} - \hat{x}_{t+1})^2$  (1)

During the learning, models have trained 10000 epochs (32 hours with Nvidia Gtx 1650 GPU/4GB Memory). 

Possible transformer usage is also tested instead of using the LSTM\cite{ichiwara2022contact} layer.

\begin{figure}
    \includegraphics[width=\linewidth]{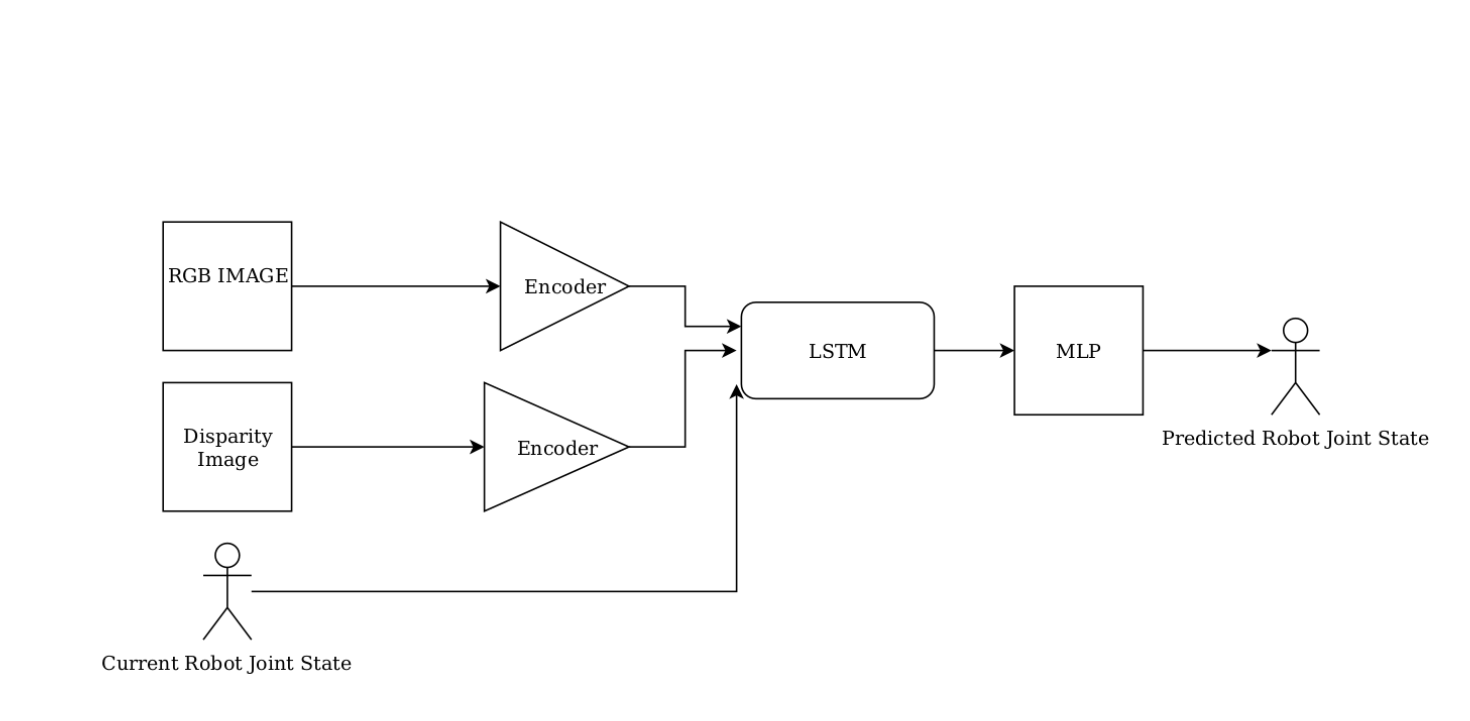}
    \caption{Neural Network Architecture }
    \label{fig:2.2}
    \end{figure}

\section{Experimental Result}
The first experiment with Encoder-LSTM-Decoder\cite{ichiwara2022contact} with long horizontal trajectories was unsuccessful. The system could not sufficiently decrease loss function, showing that the current neural network architecture cannot learn very long and diverse action patterns. Another problem related to this case is that the system has a huge output state(movebase\cite{zheng2021ros} state and all joints).

Our neural network learning using a dataset not using movebase\cite{zheng2021ros} with a short horizon successfully minimized the loss function. The experimental result is shown in figure 6. One problem related to default moving grasping makes different manipulating objects, such as approaching an object from its left or right side. Still, it does not produce extremely similar grasping patterns like teleoperation\cite{mandlekar2020human} and kinesthetic\cite{akgun2012trajectories} teaching.

\begin{figure}
    \includegraphics[width=\linewidth]{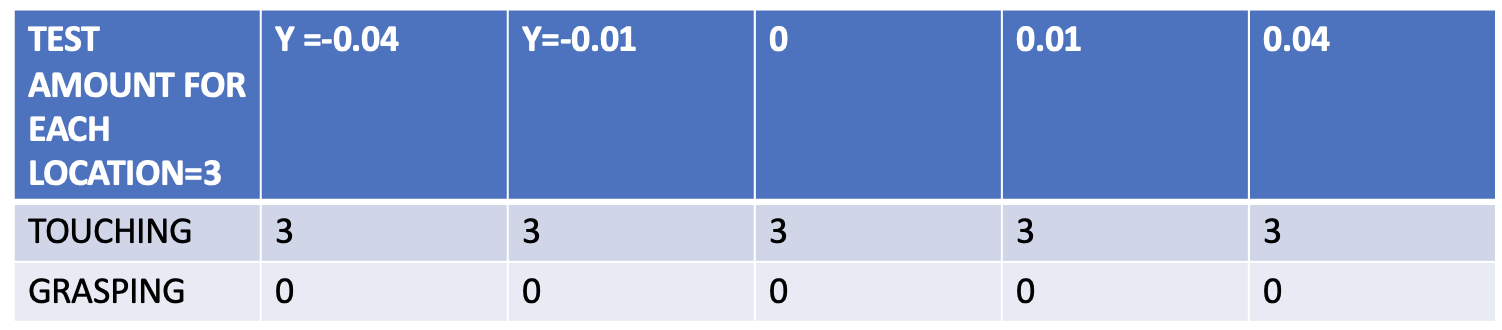}
    \caption{Experimental Result }
    \label{fig:2.2}
    \end{figure}

\begin{figure}
    \includegraphics[width=\linewidth]{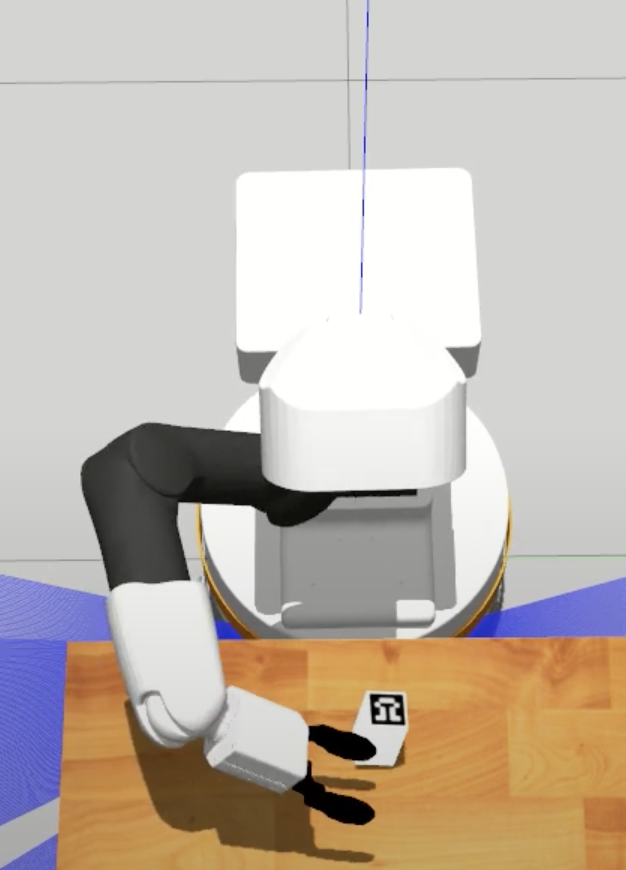}
    \caption{The moment Tiago\cite{pages2016tiago} touches the object }
    \label{fig:2.2}
    \end{figure}

Last part of the experiment, transformer usage is investigated instead of using  LSTM layers\cite{ichiwara2022contact}. It is seen that transformers\cite{dasari2021transformers}' learning has a slower learning speed compared to LSTM\cite{ichiwara2022contact}. Because of the slow learning nature of transformers\cite{dasari2021transformers}, the experiment is stopped. The result could not be investigated.

\section{Discussion}
During the experiment, it is shown that the hardware system plays the most important role. Because of hardware system limitations,  it is highly possible that our learning system for a long horizontal dataset failed. . Firstly, our current hardware did not support automated dataset production; moreover learning time of a very big dataset is extremely high for our current hardware. This situation hindered our experiment for long horizontal datasets and also big dataset collection case. In contrast to a problem faced during the first experiment, the robot could learn successfully touch the object for each training case for our limited scenario. It also shows that our system is able to learn dataset-generated classical robotics codes, and it is open to further development.

This experiment shows that robot learning tasks are generally based on similar trajectories and environments. This situation is mainly related to a limited dataset and hardware system limitations. At the same time, a not diverse dataset limits the generalistic quality of the trained neural network, and they perform overfitted tasks. This situation is undesirable for deep learning-based robotics applications in real life. Robots developed not generalistic datasets might easily fail to perform the basic task, and their domain adaptation characteristic is extremely limited.

Our experiment addressed one problem related to MoveIt\cite{chitta2016moveit} . Producing similar grasping patterns from MoveIt\cite{chitta2016moveit}  is a bit complicated. Actually, having a big diverse dataset is something desirable, but in our case, limited computational power blocks our neural networks learning performance. Firstly,  having a big dataset extremely extends our learning time, and diverse motions in datasets decrease our neural networks learning performance. It is highly expected that our mechanism might produce better results with strong high-performance computing system experiments showing the possibility of generating a dataset based on classical robotics algorithms. 
Our neural network's success on the second dataset shows the importance of classical robotics algorithms for dataset generation. Clearly, kinesthetic\cite{akgun2012trajectories} teaching,  teleoperation\cite{mandlekar2020human}, and reinforcement learning \cite{fayjie2018driverless} have massive efficiency problems. They are most likely not the optimal way for collecting robotics datasets.

\section{Future Improvement}
If the system will be developed with fewer computing units, even though it is undesirable to have a similar pattern, improving MoveIt\cite{chitta2016moveit}  motion grasping and making their grasping pattern similar is a must for increasing the success of grasping actions. In this way, our system losses its generalization ability but increase its performance in the specific grasping task.

Another alternative might be using/ action-based robotics approaches\cite{celik2022specializing}. The mixture of expert models\cite{celik2022specializing} is one of the popular ways to develop an action-based robotics system \cite{celik2022specializing}. Instead of developing end-to-end learning, developing multiple models for different actions  might increase overall performance. As an example, for dataset-1, the user can collect four different action datasets for each action, such as: approaching the object, opening the Tiago\cite{pages2016tiago} arm, grasping the object, and placing the object. Afterward can train four different models for each action, expertized on just one of them, and uses MoE\cite{celik2022specializing} for action selection. It again may lose from generalization but increase its performance on specific tasks.

Moreover,  model learning\cite{nguyen2011model} might also increase overall performance. It could be a great idea to teach two neural networks, one for just controlling robot gripper position like a cartesian controller and another one just learning robots gripper trajectories based on their location. In this way, we can reduce our big unknown joint space states and motion states to a limited number. Instead of using a learning-based cartesian controller, it could be a good idea to use some classical controller systems.

Lastly, human nature is a system that learns from its previous experience. It means that human nature learns to learn. This pattern brings the question of why not robot does not apply the same principle. Meta-learning\cite{yu2018one} is a method for developing a system based on using train data to learn. Afterward, the user feeds task data to the model, and the system performs exceptionally well in many experiments. Our classical robotics-based dataset collection is an excellent way to collect the train data, and it can also collect vast task data. This way, the system can improve its overall performance.

Combining all mentioned mechanisms might also produce some big achievements.

\section{Conclusion}

The amount of implemented robotics code (not learning-based) is high. These codes are generally appropriate for automated dataset production in simulation environments and real life. Ignoring these advanced robot codes and focusing on kinesthetic\cite{akgun2012trajectories} teaching-based or teleoperation\cite{mandlekar2020human}-based dataset collection could be more inefficient. This experiment first showed that dataset collection with classical robotics codes could contribute to the area of robot learning. It is known that some tasks are not solvable via classical robotics, but their trajectories can be recorded via teleoperations\cite{mandlekar2020human} and kinesthetic\cite{akgun2012trajectories} teaching. 

This experiment fingers the idea of classical robotics-based dataset collection for meta-learning\cite{yu2018one}. Datasets created via classical robotics can be used as meta-train datasets\cite{yu2018one}. It also addresses a lot of future development ideas based on action-based, model learning, and controller-based learning. 

This experiment also emphasizes the importance of balancing dataset diversity. Due to limited computation, a big, diverse dataset  generally damages the learning of robots.

\printbibliography
\end{document}